\documentclass[11pt, a4paper, logo, copyright, nonumbering]{deepseek}
\usepackage[authoryear, sort&compress, round]{natbib}
\usepackage{dblfloatfix}
\usepackage{ulem}
\usepackage{caption}
\usepackage{dramatist}
\usepackage{xspace}
\usepackage{pifont}
\usepackage{multirow}
\usepackage{tcolorbox}
\usepackage{xltabular}
\usepackage{longtable}
\usepackage{hyperref}
\usepackage[capitalize,noabbrev]{cleveref} 
\usepackage{lipsum}  % 用于生成示例文本
\usepackage{multicol} % 引入 multicol 宏包
\usepackage{amsfonts}
\usepackage{amsmath}
\usepackage{amssymb}
\usepackage{lineno}

\usepackage{mathtools}
\usepackage{mathrsfs}
\usepackage{physics}

\usepackage[bottom]{footmisc}
\usepackage{CJKutf8}
\usepackage{subfigure}
\usepackage{setspace}

\makeatletter
\def\@BTrule[#1]{%
  \ifx\longtable\undefined
    \let\@BTswitch\@BTnormal
  \else\ifx\hline\LT@hline
    \nobreak
    \let\@BTswitch\@BLTrule
  \else
     \let\@BTswitch\@BTnormal
  \fi\fi
  \global\@thisrulewidth=#1\relax
  \ifnum\@thisruleclass=\tw@\vskip\@aboverulesep\else
  \ifnum\@lastruleclass=\z@\vskip\@aboverulesep\else
  \ifnum\@lastruleclass=\@ne\vskip\doublerulesep\fi\fi\fi
  \@BTswitch}
\makeatother

\addto\extrasenglish{
}

 {\begin{list}{}%
         {\setlength{\leftmargin}{#1}}%
         \item[]%
 }
 {\end{list}}

\reportnumber{001} % Leave blank if n/a

\renewcommand{\today}{}
\definecolor{gold}{HTML}{D4AF37}
\def\newmodel{DeepSeek-V3.2}
\def\highmodel{DeepSeek-V3.2-Speciale}
\def\oldmodel{DeepSeek-V3.1-Terminus}
\def\methodfull{DeepSeek Sparse Attention}
\def\method{DSA}

\DeclarePairedDelimiterX{\infdivx}[2]{(}{)}{#1 \, \delimsize\| \, #2}
\newcommand{\Dkl}{\mathbb{D}_{\mathrm{KL}}\infdivx*}

\title{\centering \newmodel: Pushing the Frontier of Open \\ Large Language Models }

\author[*]{
\vspace{-0.5cm}
DeepSeek-AI
\\
\small
\texttt{research@deepseek.com}
\\
\small
\vspace{-0.5cm}
}

\renewcommand{\phi}{\varphi}

\renewcommand{\epsilon}{\varepsilon}
\renewcommand{\imath}{\mathrm{i}}

\newlength{\restsubwidth}
\newlength{\restsubheight}
\newlength{\restsubmoreheight}
\newcommand{\rest}[2]{%
        \settowidth{\restsubwidth}{\ensuremath{#2}}
        \settoheight{\restsubheight}{\ensuremath{{}_{#2}}}
        \ensuremath{{#1\hskip 0.5pt}_{\vrule\kern2pt\parbox[b][%
        4pt][b]{\the\restsubwidth}{%
                        \ensuremath{{}_{#2}}}}}
        }

\newcommand{\Jgrpo}{\mathcal{J}_{\mathrm{GRPO}}}
\newcommand{\piold}{\pi_{\mathrm{old}}}
\newcommand{\piref}{\pi_{\mathrm{ref}}}

\begin{abstract}

We introduce \newmodel, a model that harmonizes high computational efficiency with superior reasoning and agent performance. The key technical breakthroughs of \newmodel{} are as follows: \textbf{(1)  DeepSeek Sparse Attention (DSA)}:
We introduce DSA, an efficient attention mechanism that substantially reduces computational complexity while preserving model performance in long-context scenarios.
\textbf{(2) Scalable Reinforcement Learning Framework}: By implementing a robust reinforcement learning protocol and scaling post-training compute, DeepSeek-V3.2 performs comparably to GPT-5. Notably, our high-compute variant, \highmodel{}, surpasses GPT-5 and exhibits reasoning proficiency on par with Gemini-3.0-Pro, achieving \textcolor{gold}{gold-medal} performance in both the 2025 International Mathematical Olympiad (IMO) and the International Olympiad in Informatics (IOI).
\textbf{(3) Large-Scale Agentic Task Synthesis Pipeline}: To integrate reasoning into tool-use scenarios, we developed a novel synthesis pipeline that systematically generates training data at scale. This methodology facilitates scalable agentic post-training, yielding substantial improvements in generalization and instruction-following robustness within complex, interactive environments.

\end{abstract}

\begin{document}
\begin{CJK*}{UTF8}{gbsn}

\maketitle

\begin{figure}[h]
\centering
\includegraphics[width=1.0\textwidth]{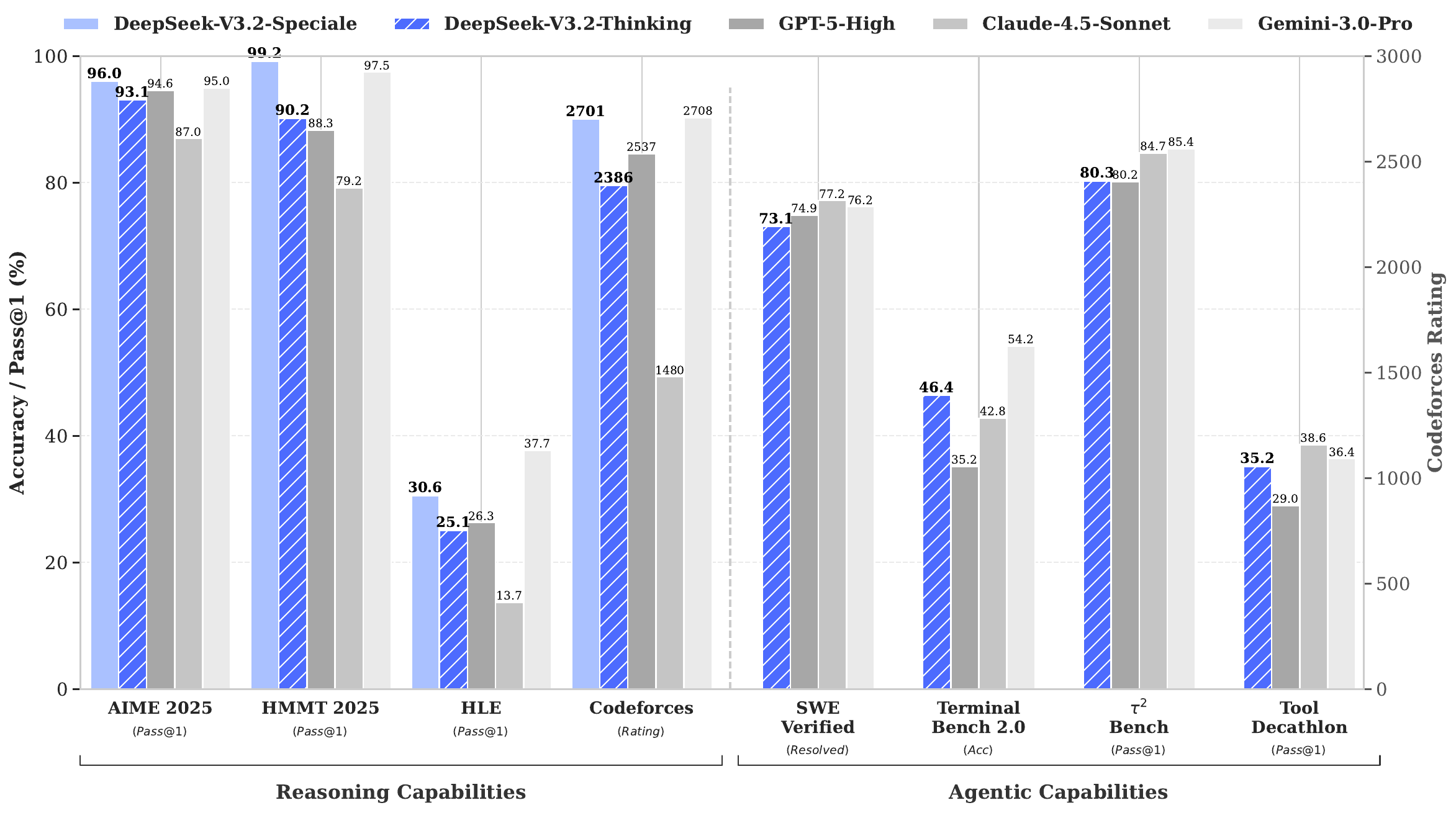}
\caption{
    \centering
    Benchmark of \newmodel{} and its counterparts. For HMMT 2025, we report the February competition, consistent with the baselines. For HLE, we report the text-only subset.
}
\label{fig:dsv3_performance}
\end{figure}

\section{Introduction}

The release of reasoning models \citep{o1, deepseekr1} marked a pivotal moment in the evolution of Large Language Models (LLMs), catalyzing a substantial leap in overall performance across the verifiable fields. Since this milestone, the capabilities of LLMs have advanced rapidly. However, a distinct divergence has emerged in the past months. While the open-source community \citep{yang2025qwen3technicalreport, zeng2025glm,MiniMax-M2, k2-thinking} continues to make strides, the performance trajectory of closed-source proprietary models \citep{gpt-5, sonnet-4.5, comanici2025gemini} has accelerated at a significantly steeper rate. Consequently, rather than converging, the performance gap between closed-source and open-source models appears to be widening, with proprietary systems demonstrating increasingly superior capabilities in complex tasks.

Through our analysis, we identify three critical deficiencies that limit the capability of open-source models in complex tasks. First, architecturally, the predominant reliance on vanilla attention \citep{vaswani2017attention} mechanisms severely constrains efficiency for long sequences. This inefficiency poses a substantial obstacle to both scalable deployment and effective post-training. Second, regarding resource allocation, open-source models suffer from insufficient computational investment during the post-training phase, limiting their performance on hard tasks. Finally, in the context of AI agents, open-source models demonstrate a marked lag in generalization and instruction-following capabilities compared to their proprietary counterparts \citep{mcpmark,mcpuniverse,li2025tool}, hindering their effectiveness in real deployment.
%, such as the International Olympiad in Informatics (IOI)\footnote{\url{https://ioinformatics.org/}} and the International Mathematical Olympiad (IMO)\footnote{\url{https://www.imo-official.org/}}

To address these critical limitations, we first introduce DSA, a highly efficient attention mechanism designed to substantially reduce computational complexity. This architecture effectively addresses the efficiency bottleneck, preserving model performance even in long-context scenarios. Second, we develop a stable and scalable RL protocol that allows for significant computational expansion during the post-training phase. Notably, this framework allocates a post-training computational budget exceeding 10\% of the pre-training cost, unlocking advanced capabilities. Thirdly, we propose a novel pipeline to foster generalizable reasoning in tool-use scenarios. First, we implement a cold-start phase utilizing the DeepSeek-V3 \citep{deepseekv3} methodology to unify reasoning and tool-use within single trajectories. Subsequently, we advance to large-scale agentic task synthesis, where we generate over 1,800 distinct environments and 85,000 complex prompts. This extensive synthesized data drives the RL process, significantly enhancing the model's generalization and instruction-following capability in the agent context. 

\newmodel{} achieves similar performance with Kimi-k2-thinking and GPT-5 across multiple reasoning benchmarks. Furthermore, \newmodel{} significantly advances the agentic capabilities of open models, demonstrating exceptional proficiency on the long-tail agent tasks introduced in \citet{mcpmark,mcpuniverse,li2025tool}. \newmodel{} emerges as a highly cost-efficient alternative in agent scenarios, significantly narrowing the performance gap between open and frontier proprietary models while incurring substantially lower costs.
Notably, with the aim of pushing the boundaries of open models in the reasoning domain, we relaxed the length constraints to develop \highmodel{}. As a result, \highmodel{} achieves performance parity with the leading closed-source system, Gemini-3.0-Pro \citep{gemini3}. It shows gold-medal performance in the IOI 2025, ICPC World Final 2025, IMO 2025, and CMO 2025.

\section{\newmodel{} Architecture}

\subsection{DeepSeek Sparse Attention}
\newmodel{} uses exactly the same architecture as DeepSeek-V3.2-Exp. 
Compared with \oldmodel{}, the last version of DeepSeek-V3.1, the only architectural modification of \newmodel{} is the introduction of \methodfull{}~(\method{}) through continued training.

\paragraph{Prototype of \method{}.}
The prototype of \method{} primarily consists of two components: a lightning indexer and a fine-grained token selection mechanism.

The \textbf{lightning indexer} computes the index score $I_{t, s}$ between the query token $\mathbf{h}_t \in \mathbb{R}^{d}$ and a preceding token $\mathbf{h}_s \in \mathbb{R}^{d}$, determining which tokens to be selected by the query token: 
\begin{equation}
    I_{t, s} = \sum_{j=1}^{H^I} w_{t, j}^I \cdot \text{ReLU}\left(\mathbf{q}^{I}_{t, j} \cdot \mathbf{k}^{I}_{s}\right),
\end{equation}
where $H^{I}$ denotes the number of indexer heads; $\mathbf{q}^{I}_{t, j} \in \mathbb{R}^{d^{I}}$ and $w_{t, j}^I \in \mathbb{R}$  are derived from the query token $\mathbf{h}_t$;
and $\mathbf{k}^{I}_{s} \in \mathbb{R}^{d^{I}}$ is derived from the preceding token $\mathbf{h}_s$. 
We choose ReLU as the activation function for throughput consideration.
Given that the lightning indexer has a small number of heads and can be implemented in FP8, its computational efficiency is remarkable. 

Given the index scores $\{I_{t, s}\}$ for each query token $\mathbf{h}_t$, our \textbf{fine-grained token selection mechanism} retrieves only the key-value entries $\{\mathbf{c}_s\}$ corresponding to the top-k index scores. 
Then, the attention output $\mathbf{u}_t$ is computed by applying the attention mechanism between the query token $\mathbf{h}_t$ and the sparsely selected key-value entries $\{\mathbf{c}_s\}$:
\begin{equation}
    \mathbf{u}_t = \text{Attn}\qty(\mathbf{h}_t, \qty{ \mathbf{c}_s \, \middle| \, I_{t, s} \in \text{Top-k} \qty(I_{t, :}) } ).
\end{equation}

\begin{figure}[t]
    \centering
    \includegraphics[width=0.95\linewidth]{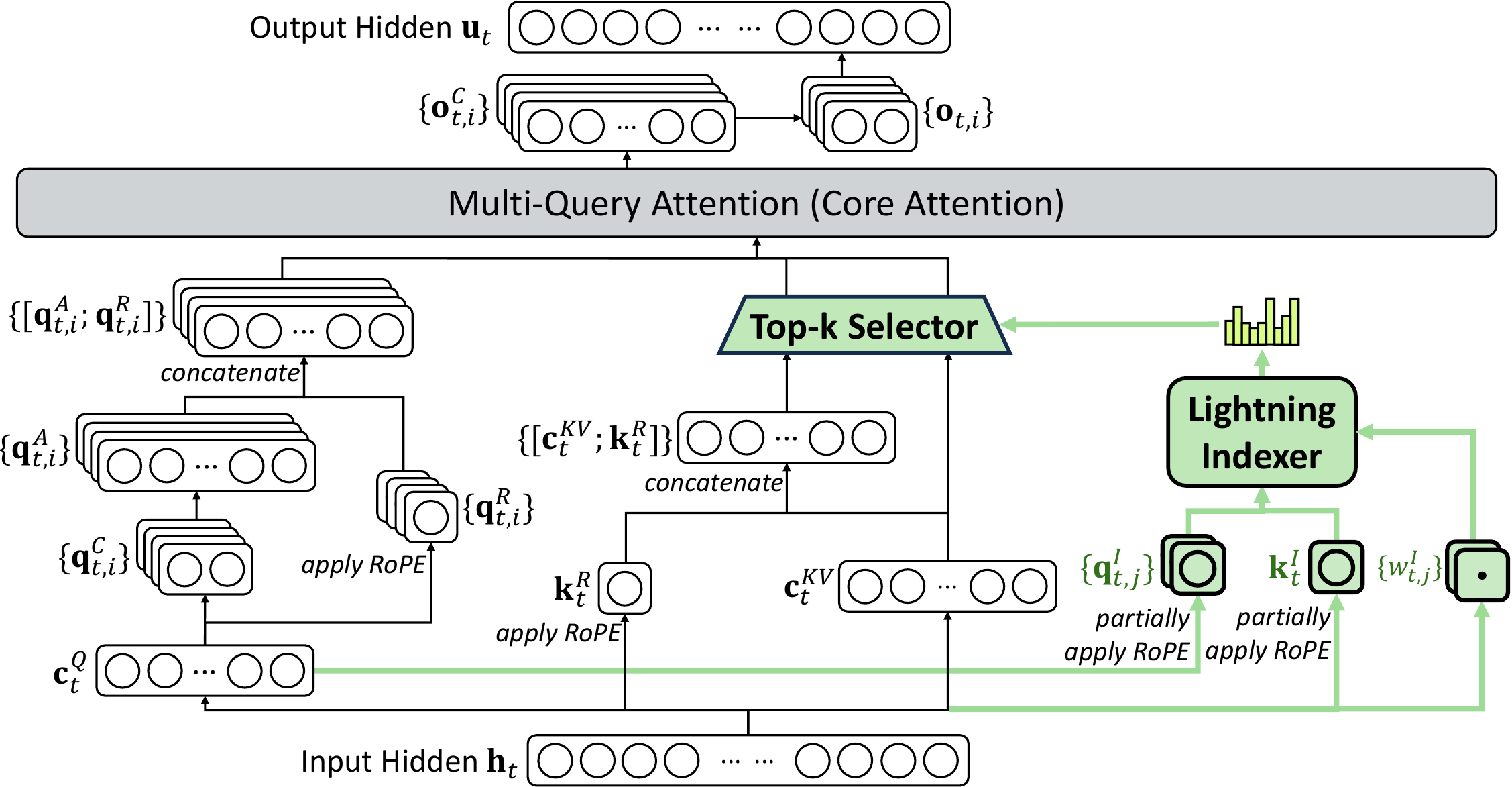}
    \caption{
    Attention architecture of \newmodel{}, where \method{} is instantiated under MLA. 
    The green part illustrates how \method{} selects the top-k key-value entries according to the indexer.
    }
    \label{fig:dsa_mla}
\end{figure}

\paragraph{Instantiate \method{} Under MLA.}
For the consideration of continued training from \oldmodel{}, we instantiate DSA based on MLA~\citep{deepseekV2} for \newmodel{}.
At the kernel level, each key-value entry must be shared across multiple queries for computational efficiency~\citep{yuan-etal-2025-native}. 
Therefore, we implement DSA based on the MQA~\citep{MQA} mode of MLA\footnote{We illustrate the difference between the MQA and MHA modes of MLA in Appendix~\ref{appendix:mqa_mha}.}, where each latent vector (the key-value entry of MLA) will be shared across all query heads of the query token. 
The DSA architecture based on MLA is illustrated in Figure~\ref{fig:dsa_mla}.
We also provide an open-source implementation of \newmodel{}\footnote{\url{https://huggingface.co/deepseek-ai/DeepSeek-V3.2-Exp/tree/main/inference}} to specify the details unambiguously.

\subsubsection{Continued Pre-Training}

Starting from a base checkpoint of \oldmodel{}, whose context length has been extended to 128K, we perform continued pre-training followed by post-training to create \newmodel{}.

The continued pre-training of \newmodel{} consists of two training stages. 
For both stages, the distribution of training data is totally aligned with the 128K long context extension data used for \oldmodel{}. 

\paragraph{Dense Warm-up Stage.}
We first use a short warm-up stage to initialize the lightning indexer. 
In this stage, we keep dense attention and freeze all model parameters except for the lightning indexer. 
To align the indexer outputs with the main attention distribution, for the $t$-th query token, we first aggregate the main attention scores by summing across all attention heads. 
This sum is then L1-normalized along the sequence dimension to produce a target distribution $p_{t,:} \in \mathbb{R}^{t}$.
Based on $p_{t,:}$, we set a KL-divergence loss as the training objective of the indexer:
\begin{equation}
    \mathcal{L}^{I} = \sum_t \Dkl{p_{t,:}}{\text{Softmax}\qty({I}_{t,:})}.
\end{equation}
For warm-up, we use a learning rate of $10^{-3}$. 
We train the indexer for only 1000 steps, with each step consisting of 16 sequences of 128K tokens, resulting in a total of 2.1B tokens.

\paragraph{Sparse Training Stage.}
Following indexer warm-up, we introduce the fine-grained token selection mechanism and optimize all model parameters to adapt the model to the sparse pattern of \method{}. 
In this stage, we also keep aligning the indexer outputs to the main attention distribution, but considering only the selected token set $\mathcal{S}_t=\qty{s \, \middle| \, I_{t,s} \in \text{Top-k} \qty(I_{t,:})}$:
\begin{equation}
    \mathcal{L}^{I} = \sum_t \Dkl{p_{t,\mathcal{S}_t}}{\text{Softmax}\qty(I_{t,\mathcal{S}_t})}. 
\end{equation}
It is worth noting that we detach the indexer input from the computational graph for separate optimization. 
The training signal of the indexer is from only $\mathcal{L}^{I}$, while the optimization of the main model is according to only the language modeling loss.
In this sparse training stage, we use a learning rate of $7.3 \times 10^{-6}$, and select 2048 key-value tokens for each query token.
We train both the main model and the indexer for $15000$ steps, with each step consisting of 480 sequences of 128K tokens, resulting in a total of 943.7B tokens.

\subsection{Parity Evaluation}

\paragraph{Standard Benchmark}
In September 2025, we evaluate DeepSeek-V3.2-Exp on a suite of benchmarks, which focus on diverse capabilities, and compare it with \oldmodel{} showing similar performance. 
While DeepSeek V3.2 Exp significantly improves computational efficiency on long sequences, we do not observe substantial performance degradation compared with \oldmodel{}, on both short- and long-context tasks. 

\paragraph{Human Preference} Given that direct human preference assessments are inherently susceptible to bias, we employ ChatbotArena as an indirect evaluation framework to approximate user preferences for the newly developed base models. Both DeepSeek‑V3.1‑Terminus and DeepSeek‑V3.2‑Exp share an identical post‑training strategy, and their Elo scores, obtained from evaluations conducted on 10 November 2025, are closely matched. These results suggest that the new base model achieves performance on par with the previous iteration, despite incorporating a sparse attention mechanism.
\paragraph{Long Context Eval} Following the release of DeepSeek‑V3.2‑Exp, several independent long‑context evaluations were conducted using previously unseen test sets. A representative benchmark is AA‑LCR\footnote{\url{https://artificialanalysis.ai/evaluations/artificial-analysis-long-context-reasoning}}, in which DeepSeek‑V3.2‑Exp scores four points higher than \oldmodel~ in reasoning mode. In the Fiction.liveBench evaluation\footnote{\url{https://fiction.live/stories/Fiction-liveBench-April-6-2025/oQdzQvKHw8JyXbN87}}, DeepSeek‑V3.2‑Exp consistently outperforms \oldmodel~ across multiple metrics. This evidence indicates the base checkpoint of DeepSeek‑V3.2‑Exp does not regress on long context tasks.

\subsection{Inference Costs}
\begin{figure}[t]
    \centering
    \subfigure[Prefilling]{
        \includegraphics[width=0.475\textwidth]{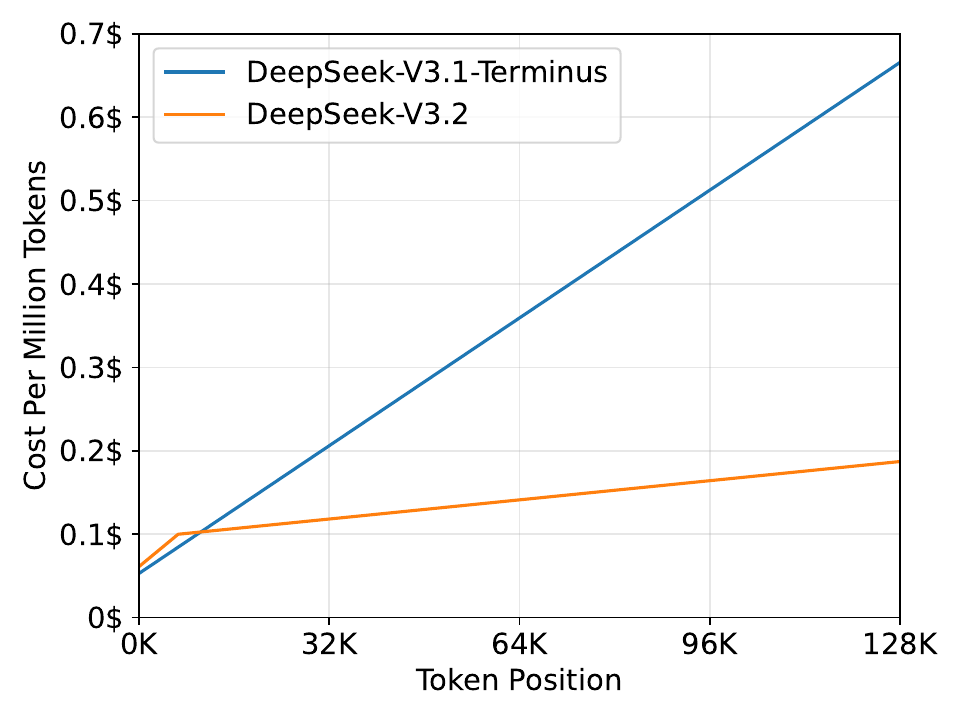}
        \label{fig:cost_prefilling}
    }
    \hspace{0.01cm}
    \subfigure[Decoding]{
        \includegraphics[width=0.475\textwidth]{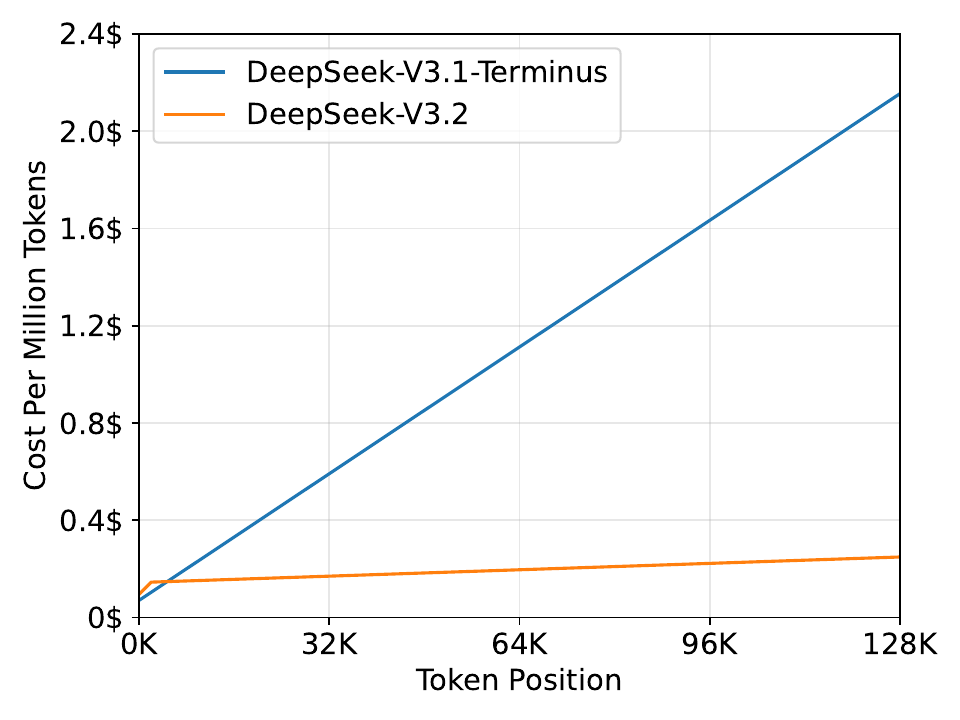}
        \label{fig:cost_decoding}
    }
    \caption{
    Inference costs of \oldmodel{} and \newmodel{} on H800 clusters.
    }
    \label{fig:cost}
\end{figure}

\method{} reduces the core attention complexity of the main model from $\order{L^2}$ to $\order{L k}$, where $k$ ($\ll L$) is the number of selected tokens. 
Although the lightning indexer still has a complexity of $\order{L^2}$, it requires much less computation compared with MLA in \oldmodel{}.
Combined with our optimized implementation, \method{} achieves a significant end-to-end speedup in long-context scenarios.
Figure~\ref{fig:cost} presents how token costs of \oldmodel{} and \newmodel{} vary with the token position in the sequence. 
These costs are estimated from benchmarking the actual service deployed on H800 GPUs, at a rental price of 2 USD per GPU hour.
Note that for short-sequence prefilling, we specially implement a masked MHA mode to simulate \method{}, which can achieve higher efficiency under short-context conditions.

\section{Post-Training}

After continued pre-training, we perform post-training to create the final \newmodel{}.
The post-training of \newmodel{} also employs sparse attention in the same way as the sparse continued pre-training stage. 
For \newmodel{}, we maintain the same post-training pipeline as in DeepSeek-V3.2-Exp, which includes specialist distillation and mixed RL training. 

\paragraph{Specialist Distillation} For each task, we initially develop a specialized model dedicated exclusively to that particular domain, with all specialist models being fine-tuned from the same pre-trained DeepSeek-V3.2 base checkpoint. 
In addition to writing tasks and general question-answering, our framework encompasses six specialized domains: mathematics, programming, general logical reasoning, general agentic tasks, agentic coding, and agentic search, with all the domains supporting both thinking and non-thinking modes. 
Each specialist is trained with large-scale Reinforcement Learning (RL) computing.
Furthermore, we employ different models to generate training data for long chain-of-thought reasoning (thinking mode) and direct response generation (non-thinking mode). 
Once the specialist models are prepared, they are used to produce the domain-specific data for the final checkpoint. 
Experimental results demonstrate that models trained on the distilled data achieve performance levels only marginally below those of domain-specific specialists, with the performance gap being effectively eliminated through subsequent RL training.

\paragraph{Mixed RL Training} For \newmodel{}, we still adopt Group Relative Policy Optimization (GRPO)~\citep{deepseekmath,deepseekr1} as the RL training algorithm. 
As DeepSeek-V3.2-Exp, we merge reasoning, agent, and human alignment training into one RL stage. 
This approach effectively balances performance across diverse domains while circumventing the catastrophic forgetting issues commonly associated with multi-stage training paradigms. 
For reasoning and agent tasks, we employ rule-based outcome reward, length penalty, and language consistency reward. 
For general tasks, we employ a generative reward model where each prompt has its own rubrics for evaluation.

\paragraph{\newmodel{} and \highmodel{}}
\newmodel{} integrates reasoning, agent, and human alignment data distilled from specialists, undergoing thousands of steps of continued RL training to reach the final checkpoints.
To investigate the potential of extended thinking, we also developed an experimental variant, \highmodel{}. This model was trained exclusively on reasoning data with a reduced length penalty during RL. Additionally, we incorporated the dataset and reward method from DeepSeekMath-V2 \citep{deepseek-math-v2} to enhance capabilities in mathematical proofs.

We would like to highlight our efforts in how to create a stable recipe to scale up RL compute in Section \ref{sec:grpo}, and how to integrate thinking into agentic tasks in Section \ref{sec:synthesis}

\subsection{Scaling GRPO}
\label{sec:grpo}

%\paragraph{RL Algorithm.} %We adopt Group Relative Policy Optimization (GRPO) in the RL training stage[cite DS-Math; cite R1].
We first review the objective of GRPO. GRPO optimizes the policy model $\pi_{\theta}$ by maximizing the following objective on a group of responses $\{o_1,\cdots, o_G\}$ sampled from the old policy $\piold$ given each question $q$:
\begin{align}
    \Jgrpo(\theta)=\enspace& \mathbb{E}_{
        q\sim P(Q), 
        \{o_i\}_{i=1}^G\sim \piold(\cdot|q)}\Bigg[
            \frac{1}{G}\sum_{i=1}^G\frac{1}{|o_i|}\sum_{t=1}^{|o_i|} \nonumber\\
            &\min\left(
                r_{i,t}(\theta) \hat{A}_{i,t}, 
                \text{clip}\left(r_{i,t}(\theta), 1-\varepsilon, 1+\varepsilon\right)\hat{A}_{i,t}
            \right) - \beta \Dkl{\pi_\theta(o_{i,t})}{\piref(o_{i,t})}
        \Bigg],
\end{align}
where
\begin{equation}
    r_{i,t}(\theta)=\frac{\pi_\theta(o_{i,t}|q,o_{i,<t})}{\piold(o_{i,t}|q,o_{i,<t})}
\end{equation}
is the importance sampling ratio between the current and old policy. $\varepsilon$ and $\beta$ are hyper-parameters controlling the clipping range and KL penalty strength, respectively. $\hat{A}_{i,t}$ is the advantage of $o_{i,t}$ which is estimated by normalizing the outcome reward within each group. Specifically, a set of reward models are used to score an outcome reward $R_i$ for each output $o_i$ in the group, yielding $G$ rewards $\boldsymbol{R}=\{R_1,\cdots,R_G\}$ respectively. The advantage of $o_{i,t}$ is calculated by subtracting the average reward of the group from the reward of output $o_i$, i.e., $\hat{A}_{i,t} = R_i-\text{mean}(\boldsymbol{R})$. 

In the following, we outline additional strategies that stabilize RL scaling, directly building on the GRPO algorithm.

\paragraph{Unbiased KL Estimate} Given $o_{i,t}$ is sampled from the old policy $\piold(\cdot|q,o_{i,<t})$, we correct the K3 estimator \citep{schulman2020klapprox} to obtain an unbiased KL estimate
using the importance-sampling ratio between the current policy $\pi_\theta$ and the old policy $\piold$ .
\begin{equation}
    \Dkl{\pi_\theta(o_{i,t})}{\piref(o_{i,t})} = 
        \frac{\pi_\theta(o_{i,t}|q,o_{i,<t})}{\piold(o_{i,t}|q,o_{i,<t})}
        \left(
            \frac{\piref(o_{i,t}|q,o_{i,<t})}{\pi_{\theta}(o_{i,t}|q,o_{i,<t})}
            - \log \frac{\piref(o_{i,t}|q,o_{i,<t})}{\pi_{\theta}(o_{i,t}|q,o_{i,<t})} - 1
        \right).
\end{equation}

As a direct result of this adjustment, the gradient of this KL estimator becomes unbiased, which eliminates systematic estimation errors, thereby facilitating stable convergence. This contrasts sharply with the original K3 estimator, particularly when the sampled tokens have substantially lower probabilities under the current policy than the reference policy, i.e., $\pi_\theta \ll \piref$. In such cases, the gradient of the K3 estimator assigns disproportionately large, unbounded weights to maximize the likelihood of these tokens, resulting in noisy gradient updates that accumulate to degrade sample quality in subsequent iterations and lead to unstable training dynamics. In practice, we find that different domains benefit from varying strengths of KL regularization. For certain domains, such as mathematics, applying a relatively weak KL penalty or even omitting it entirely can yield improved performance.

\paragraph{Off-Policy Sequence Masking} To improve the efficiency of RL systems, we typically generate a large batch of rollout data, which is subsequently split into multiple mini-batches for several gradient update steps. This practice inherently introduces off-policy behavior. Additionally, inference frameworks used for efficient data generation are often highly optimized, which may differ in implementation details from training frameworks. Such training-inference inconsistency further exacerbates the degree of off-policyness. To stabilize training and improve tolerance for off-policy updates, we mask negative sequences that introduce significant policy divergence, as measured by the KL divergence between the data-sampling policy $\piold$ and the current policy $\pi_\theta$. More specifically, we introduce a binary mask $M$ into the GRPO loss:
\begin{align}
    \Jgrpo(\theta)=\enspace& \mathbb{E}_{
        q\sim P(Q), 
        \{o_i\}_{i=1}^G\sim \piold(\cdot|q)}\Bigg[
            \frac{1}{G}\sum_{i=1}^G\frac{1}{|o_i|}\sum_{t=1}^{|o_i|} \nonumber\\
            &\min\left(
                r_{i,t}(\theta) \hat{A}_{i,t}, 
                \text{clip}\left(r_{i,t}(\theta), 1-\varepsilon, 1+\varepsilon\right)\hat{A}_{i,t}
            \right)M_{i,t} - \beta \Dkl{\pi_\theta(o_{i,t})}{\piref(o_{i,t})}
        \Bigg],
\end{align}
where

\begin{equation}
M_{i,t} = \begin{cases}
0 & {\hat{A}_{i,t} < 0, \frac{1}{|o_i|}\sum_{t=1}^{|o_i|}\log\frac{\piold(o_{i,t}|q,o_{i,<t})}{\pi_{\theta}(o_{i,t}|q,o_{i,<t})} > \delta} \\[1ex]
1 & {\text{otherwise},}
\end{cases}
\end{equation}
and $\delta$ is a hyper-parameter that controls the threshold of policy divergence. Note that $\piold$ here denotes the sampling probability directly returned by the inference framework, thus the KL divergence between the old and current policy accounts for both sources of off-policyness mentioned above. It is also worth noting that we only mask sequences with negative advantages. 

Intuitively, the model benefits the most by learning from its own mistakes, whereas highly off-policy negative samples can be detrimental, potentially misleading or destabilizing the optimization process. We empirically observe that this Off-Policy Sequence Masking operation improves stability in certain training scenarios that would otherwise exhibit instability.

\paragraph{Keep Routing} Mixture-of-Experts (MoE) models improve computational efficiency by activating only a subset of expert modules during inference. However, discrepancies between inference and training frameworks, compounded by policy updates, can result in inconsistent expert routing during inference and training even for identical inputs. Such inconsistency induces abrupt shifts in the active parameter subspace, which destabilizes optimization and exacerbates off-policy issues. To mitigate this, we preserve the expert routing paths used during sampling in the inference framework and enforce the same routing paths during training, ensuring that identical expert parameters are optimized. This Keep Routing operation was found crucial for RL training stability of MoE models, and has been adopted in our RL training pipeline since DeepSeek-V3-0324.

\paragraph{Keep Sampling Mask} Top-p and top-k sampling are widely used sampling strategies to enhance the quality of responses generated by LLMs. Employing these strategies in RL training is also advantageous, as it avoids sampling extremely low-probability tokens that would be used as optimization targets. While such truncation preserves sample quality, it introduces a mismatch between the action spaces of $\piold$ and $\pi_\theta$, which violates the principles of importance sampling and destabilizes training. To address this, we preserve the truncation masks during sampling from $\piold$ and apply them to $\pi_\theta$ during training, ensuring both policies share identical action subspaces. Empirically, we find that combining top-p sampling with the Keep Sampling Mask strategy effectively preserves language consistency during RL training.

\subsection{Thinking in Tool-Use}
\label{sec:synthesis}

\subsubsection{Thinking Context Management}
DeepSeek-R1 has demonstrated that incorporating a thinking process can significantly enhance a model's ability to solve complex problems. Building on this insight, we aim to integrate thinking capabilities into tool-calling scenarios.

\begin{figure}[ht]
    \centering
    \includegraphics[width=0.95\linewidth]{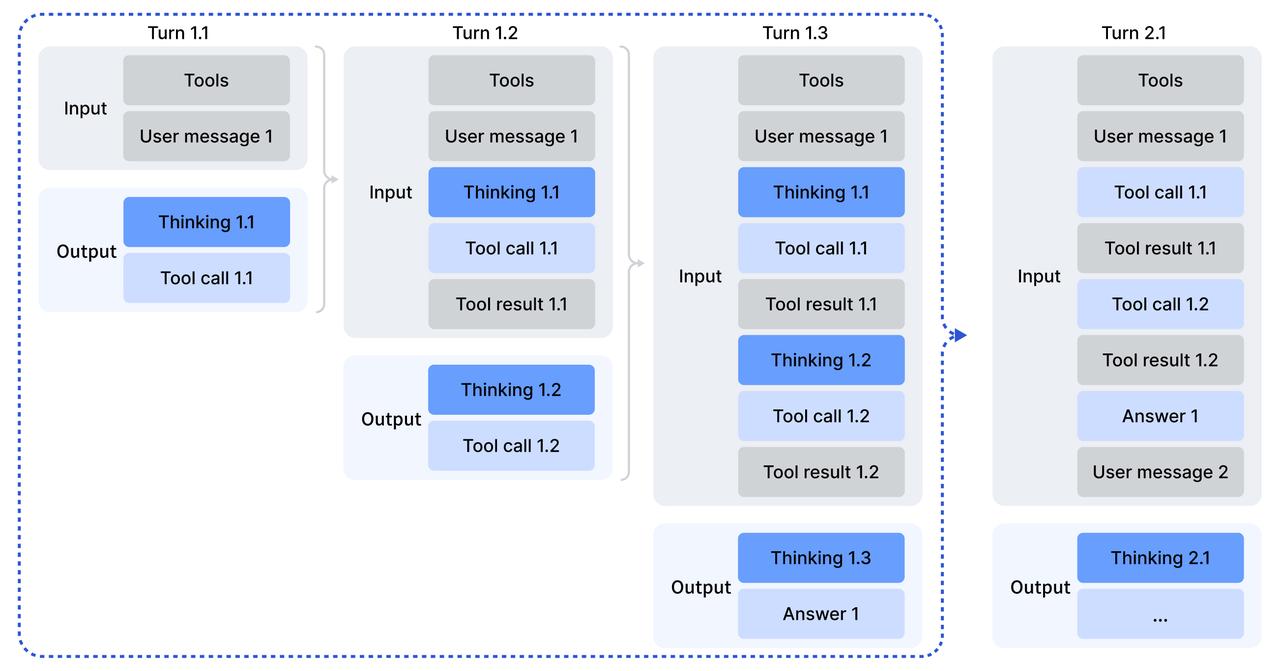}
    \caption{
        Thinking retention mechanism in tool-calling scenarios.
    }
    \label{fig:format}
\end{figure}

We observed that replicating DeepSeek-R1’s strategy—discarding reasoning content upon the arrival of the second round of messages—results in significant token inefficiency. This approach forces the model to redundantly re-reason through the entire problem for each subsequent tool call. To mitigate this, we developed a context management strictly tailored for tool-calling scenarios as shown in Fig~\ref{fig:format}:
\begin{itemize}
\item Historical reasoning content is discarded only when a new \textbf{user message} is introduced to the conversation. If only tool-related messages (e.g., tool outputs) are appended, the reasoning content is \textbf{retained} throughout the interaction.
\item When reasoning traces are removed, the history of \textbf{tool calls and their results} remains preserved in the context.
\end{itemize}
Notably, certain agent frameworks, such as Roo Code or Terminus, simulate tool interactions via user messages. These frameworks may not fully benefit from our enhanced reasoning persistence due to the context management rules outlined above. Therefore, we recommend utilizing non-thinking models for optimal performance with such architectures.

\subsubsection{Cold-Start}

Given the availability of reasoning data (non-agentic) and non-reasoning agentic data, a straightforward strategy for integrating these two capabilities is through carefully designed prompting. We posit that the model possesses sufficient ability to accurately follow explicit instructions, thereby enabling the seamless incorporation of tool execution within the reasoning process.

To demonstrate the operation of the cold-start mechanism, we selectively sample the training data as shown in Appendix Tables \ref{tab:think_template}–\ref{tab:agentthink_template}. It is important to note that distinct task prompts are associated with different system prompts. Tables \ref{tab:think_template}–\ref{tab:agentthink_template} present an illustrative example corresponding to a competitive programming prompt.
Table \ref{tab:think_template} presents an example of our reasoning data, which uses a system prompt to explicitly asks the model to do reasoning before the final answer and uses a special tag <think></think> to label the reasoning path. Table \ref{tab:agent_template} shows the prompt of non-reasoning agentic data, where the system prompt contains the guidance of toolcall. Table \ref{tab:agentthink_template} presents the system prompt we designed to instruct the model to incorporate multiple tool calls within its reasoning process.

In this manner, although the reasoning in tool‑use patterns may lack robustness, the model is occasionally able to generate the desired trajectories, thereby providing a basis for subsequent reinforcement learning stages.

\subsubsection{Large-Scale Agentic Tasks}

A diverse set of RL tasks is crucial for enhancing model robustness. For tasks such as search, code engineering, and code interpretation, we employ real-world tools, including actual web search APIs, coding tools, and Jupyter Notebooks. While these RL environments are real, the prompts employed are either extracted from Internet sources or synthetically generated, rather than obtained from actual user interactions.
For other tasks,  the environment and prompts are both synthetically constructed. The agent tasks we used are described in Table \ref{tab:description}.

\begin{table}[h]
\centering
\caption{The description of different agent tasks, including the number of tasks, environment type (real or synthesized), and prompt source (extracted or synthesized). \label{tab:description} } 
\begin{tabular}{c|c|c|c}
\hline
 & number of tasks & environment & prompt \\
\hline
code agent & 24667 & real & extracted \\
\hline
search agent& 50275 & real & synthesized \\
\hline
general agent & 4417& synthesized & synthesized\\
\hline
code interpreter & 5908 & real & extracted \\
\hline
\end{tabular}

\end{table}

\paragraph{Search Agent}

We employ a multi-agent pipeline based on DeepSeek-V3.2 to generate diverse, high-quality training data. 
We first sample informative long-tail entities across diverse domains from large-scale web corpora. 
A question-construction agent then explores each entity using search tools with configurable depth and breadth parameters, consolidating the discovered information into question-answer pairs. 
Multiple answer-generation agents with heterogeneous configurations (different checkpoints, system prompts, etc.) produce diverse candidate responses for each proposed QA pair. 
A verification agent with search capabilities validates all answers through multiple passes, retaining only samples where the ground-truth is correct and all candidates are verifiably incorrect. 
These data spans multiple languages, domains, and difficulty levels. 
To complement these verifiable samples and better reflect real-world usage, we also augment the dataset with filtered instances from our existing helpful RL datasets, for which the search tool provides measurable benefits.
We then develop detailed evaluation rubrics across multiple quality dimensions and employ a generative reward model to score responses based on these rubrics. 
This hybrid approach enables optimization for both factual reliability and practical helpfulness.

\paragraph{Code Agent}

We constructed large-scale, executable environments for software issue resolution by mining millions of issue-Pull Request (PR) pairs from GitHub. This dataset was rigorously filtered using heuristic rules and LLM-based judgments to ensure high quality, requiring that each entry contain a reasonable issue description, a correlated gold patch, and a test patch for validation. An automated environment-setup agent, powered by DeepSeek-V3.2, was employed to build executable environments for these pairs. This agent handles package installation, dependency resolution, and test execution. Test results are output in the standard JUnit format, ensuring consistent parsing across programming languages and test frameworks. An environment is deemed successfully built only when applying the gold patch results in a non-zero count of false-to-positive (F2P) test cases (indicating the issue is fixed) and a zero count of pass-to-fail (P2F) test cases (indicating no regressions). Using this pipeline, we successfully built tens of thousands of reproducible issue resolution environments spanning multiple programming languages, including Python, Java, JavaScript, TypeScript, C, C++, Go, and PHP.

\paragraph{Code Interpreter Agent}
We utilize Jupyter Notebook as a code interpreter to address complex reasoning tasks. To facilitate this, we curate a diverse set of problems spanning mathematics, logic, and data science, each requiring the model to leverage code execution capabilities to arrive at a solution.

\paragraph{General Agent}
To scale up agent environments and tasks in RL, we employ an automatic environment-synthesis agent that synthesizes 1,827 task-oriented environments. These tasks are hard to solve but easy to verify. The synthesis workflow primarily consists of environment and toolset construction, task synthesis, and solution generation. Specifically, the workflow proceeds as follows.
\begin{enumerate}
    \item Given a task category (e.g., planning a travel itinerary) and a sandbox equipped with a bash and a search tool, the agent first uses these tools to generate or retrieve relevant data from the Internet and store them in the sandbox database.
    \item The agent then synthesizes a set of task-specific tools, each implemented as a function.
    \item To create tasks that are both challenging and automatically verifiable, the agent initially proposes a simple task based on the current database, along with its solution and verification functions implemented in Python. The solution function is restricted to invoking tool functions or performing logical computations, and cannot call other functions or directly access the database, ensuring the task can only be solved through the tool interface. Additionally, the results produced by the solution function must be validated by the verification function. If the solution is not validated, the agent will modify the solution or verification functions until the solution's output passes the verification. The agent then iteratively increases the difficulty of the task and updates the corresponding solution and verification functions. During this iterative process, if the current toolset is not sufficient to solve the task, the agent will augment the toolset.

\end{enumerate}

Following this workflow, we obtain thousands of $⟨\text{environment}, \text{tools}, \text{task}, \text{verifier}⟩$ tuples. We then perform RL on this dataset using \newmodel{} and retain only instances with non-zero pass@100, resulting in 1,827 environments and their corresponding tasks (4,417 in total). A synthetic trip-planning example is illustrated below. This example highlights that, while searching the large combinatorial space for a trip plan that satisfies all constraints is challenging, checking whether a given candidate solution satisfies these constraints is relatively straightforward.
\begin{tcolorbox}[
  floatplacement=ht,    
  title={An Example of Synthesized Task: Trip Planning}, 
  label={case:trip_request}             
]
\small
I'm planning a three-day trip starting from Hangzhou, and I need help creating an itinerary from October 1st to October 3rd, 2025. 
A few important requirements: I don't want to repeat any cities, hotels, attractions, or restaurants during the entire trip. Also, please make sure that every hotel, restaurant, and attraction you recommend is actually located in the city where I'll be staying that day.
One more thing about the second day - I'm trying to be smart about my budget. If I end up booking a luxury hotel that costs 800 CNY or more per night, then I need to be more careful with other expenses: my total spending on both restaurants (lunch and dinner) should stay under 350 CNY, both restaurants should be rated at least 4.0 stars, and the afternoon attraction ticket needs to be less than 120 CNY.
If the hotel on day 2 is in the mid-to-high range (500-800 CNY), then I have a bit more flexibility - I just need to make sure at least one of my restaurant choices is rated 4.0 or higher, and the attraction ticket should be below 180 CNY.
For more affordable hotels (200-500 CNY range), I only need to ensure that at least one restaurant has a rating of 3.2 or above.
Can you help me put together this itinerary? \\[0.2em]

\textbf{Submit Result Format} \\[0.2em]

[
\\[0.2em]
  \{
    "time": "2025-10-01",
    "city": "cite\_name",
    "hotel": "hotel\_name",
    "afternoon\_restaurant": "restaurant\_name",
    "afternoon\_attraction": "attraction\_name",
    "evening\_restaurant": "restaurant\_name"
  \},
  \\[0.2em]
  \{
    "time": "2025-10-02",
    "city": "cite\_name",
    "hotel": "hotel\_name",
    "afternoon\_restaurant": "restaurant\_name",
    "afternoon\_attraction": "attraction\_name",
    "evening\_restaurant": "restaurant\_name"
  \},
  \\[0.2em]
  \{
    "time": "2025-10-03",
    "city": "cite\_name",
    "hotel": "hotel\_name",
    "afternoon\_restaurant": "restaurant\_name",
    "afternoon\_attraction": "attraction\_name",
    "evening\_restaurant": "restaurant\_name"
  \}
]
\end{tcolorbox}

\begin{tcolorbox}[
  floatplacement=ht,
  title={Tool Set for Trip Planning},
  label={case:tools}
]
\scriptsize
\begin{tabularx}{\textwidth}{l X}
\toprule
\textbf{Function Name} & \textbf{Description} \\
\midrule
\texttt{get\_all\_attractions\_by\_city(city)} &
Get all attractions for given city. \\[0.3em]

\texttt{get\_all\_cities()} &
Get all cities from the database. \\[0.3em]

\texttt{get\_all\_hotels\_by\_city(city)} &
Get all hotels for given city. \\[0.3em]

\texttt{get\_all\_restaurants\_by\_city(city)} &
Get all restaurants for given city. \\[0.3em]

\texttt{get\_city\_by\_attraction(attraction)} &
Get city for given attraction name. \\[0.3em]

\texttt{get\_city\_by\_hotel(hotel)} &
Get city for given hotel name. \\[0.3em]

\texttt{get\_city\_by\_restaurant(restaurant)} &
Get city for given restaurant name. \\[0.3em]

\texttt{get\_city\_transport(city)} &
Get all intra-city transport options for given city. \\[0.3em]

\texttt{get\_infos\_by\_attraction(info\_keywords, attraction)} &
Get specified infos for given attraction. \\[0.3em]

\texttt{get\_infos\_by\_city(info\_keywords, city)} &
Get specified infos for given city. \\[0.3em]

\texttt{get\_infos\_by\_hotel(info\_keywords, hotel)} &
Get specified infos for given hotel. \\[0.3em]

\texttt{get\_infos\_by\_restaurant(info\_keywords, restaurant)} &
Get specified infos for given restaurant. \\[0.3em]

\texttt{get\_inter\_city\_transport(from\_city, to\_city)} &
Get all transports between given city pair. \\[0.3em]

\texttt{get\_weather\_by\_city\_date(city, date)} &
Get weather for given city-date pair. \\[0.3em]

\texttt{submit\_result(answer\_text)} &
Submit the final answer content. \\
\bottomrule
\end{tabularx}
\end{tcolorbox}

% \highmodel{} is a reasoning-oriented model, which built upon an intermediate checkpoint of \newmodel{} by increasing output length. \highmodel{} does not support tool call and is inferior to \newmodel{} on general chat. During the  RL training of \highmodel{}, it applies the technique of DeepSeekMath-V2. 
% \newmodel{} does not apply RL technique in  DeepSeekMath-V2 to improve math proofing capability. 

\section{Evaluation}

\subsection{Main Results}
We evaluate models on MMLU-Pro \citep{mmlu_pro}, GPQA Diamond \citep{gpqa}, Human Last Exam (HLE) Text-only \citep{hle}, LiveCodeBench (2024.08-2025.04), Codeforces, Aider-Polyglot, AIME 2025, HMMT Feb 2025, HMMT Nov 2025 \citep{balunovic2025matharena}, IMOAnswerBench \citep{luong-etal-2025-towards}, Terminal Bench 2.0, SWE-Verified \citep{swe_verified}, SWE Multilingual \citep{yang2025swesmith}, BrowseComp \citep{wei2025browsecomp}, BrowseCompZh \citep{zhou2025browsecomp}, $\tau^2$-bench \citep{tau2}, MCP-Universe \citep{mcpuniverse}, MCP-Mark \citep{mcpmark}, and Tool-Decathlon \citep{li2025tool}. Tool-use benchmarks are evaluated using the standard function call format, wherein models are configured to thinking mode.
 For MCP-Universe \citep{mcpuniverse} and MCP-Mark \citep{mcpmark}, we evaluate all models with our internal environment, because the search and playwright environment might be slightly different from the official setting.   
We set the temperature to 1.0, and the context window to 128K tokens.
For math-related tasks such as AIME, HMMT, IMOAnswerBench, and HLE, we eval with the following template: \texttt{"\{question\}\textbackslash nPlease reason step by step, and put your final answer within \textbackslash boxed\{\}."}
In the case of HLE, we additionally assessed \newmodel{}-Thinking using the official template, resulting in a score of $23.9$.

\begin{table}[htbp]
    \centering
    \footnotesize
    \setlength{\tabcolsep}{1.9pt}
    \caption{ Comparison between \newmodel{} and closed/open models. For open models, we just compare with models supports thinking in tooluse. 
    Numbers in bold represent the best scores within each model class (open-source and closed-source).  The $\tau^2$-Bench result is computed by the average of each category. Regarding BrowseComp, the performance with the context management technique is noted with *.
      }
    \begin{tabular}{@{}c l | c  c  c | c c c c @{}}
    \toprule
    & \multirow{2}{*}{\centering \textbf{Benchmark {\tiny (Metric)}}}  & \textbf{Claude-4.5-}  & \textbf{GPT-5}& \textbf{Gemini-3.0} & \textbf{Kimi-K2} & \textbf{MiniMax} & \textbf{\newmodel} \\
    & & \textbf{Sonnet}  & \textbf{High} & \textbf{Pro} & \textbf{Thinking}& \textbf{M2} &\textbf{Thinking} \\

    \midrule
    \multirow{3}{*}{English}
    & MMLU-Pro {\tiny (EM)} & 88.2 & 87.5 & \textbf{90.1} & 84.6 & 82.0 & \textbf{85.0}\\

    & GPQA Diamond {\tiny (Pass@1)} &83.4 & 85.7& \textbf{91.9} & \textbf{84.5} & 77.7 &82.4\\
    & HLE {\tiny (Pass@1)} &13.7 & 26.3 & \textbf{37.7} & 23.9 & 12.5 & \textbf {25.1}\\
    \midrule
    \multirow{2}{*}{Code} & LiveCodeBench {\tiny (Pass@1-COT)}  & 64.0& 84.5& \textbf{90.7}& 82.6& 83.0&   \textbf{83.3} \\
    & Codeforces {\tiny (Rating)} & 1480 &2537& \textbf{2708}& -& -&2386 \\
    % & Aider-Polyglot {\tiny (Acc.)} & & & & & & \\
    \midrule
        \multirow{4}{*}{Math} & AIME 2025 {\tiny (Pass@1)}  & 87.0 & 94.6 & \textbf{95.0} & \textbf{94.5} & 78.3 & 93.1 \\
    & HMMT Feb 2025 {\tiny (Pass@1)}  & 79.2 & 88.3 & \textbf{97.5} & 89.4 & - &\textbf{92.5} \\
    & HMMT Nov 2025 {\tiny (Pass@1)}  & 81.7 & 89.2 & \textbf{93.3} & 89.2 & - &  \textbf{90.2}\\
    & IMOAnswerBench {\tiny (Pass@1)}  & - & 76.0 & \textbf{83.3} & \textbf{78.6} & -  &78.3  \\
    \midrule
     \multirow{3}{*}{Code Agent} & Terminal Bench 2.0 {\tiny (Acc)}  &42.8& 35.2& \textbf{54.2} & 35.7& 30.0 & \textbf{46.4}\\
      & SWE Verified {\tiny (Resolved)}  & \textbf{77.2} & 74.9&76.2 & 71.3& 69.4 &  \textbf{73.1}\\
      & SWE Multilingual {\tiny (Resolved)}  &\textbf{68.0} &55.3 &- & 61.1 & 56.5	 & \textbf{70.2}\\  \midrule

    \multirow{3}{*}{Search Agent} & BrowseComp {\tiny (Pass@1)}  & 24.1 & \textbf{54.9} & - & -/60.2* & 44.0& \textbf{51.4/67.6}* \\
    & BrowseCompZh {\tiny (Pass@1)}  & 42.4 & 63.0 & - & 62.3 & 48.5 & \textbf{65.0} \\
    & HLE {\tiny (Pass@1)}  &32.0 & 35.2 & \textbf{45.8} & \textbf{44.9} & 31.8 & 40.8 \\ \midrule

         \multirow{4}{*}{ToolUse}  &$\tau^2$-Bench{\tiny (Pass@1)} & 84.7& 80.2& \textbf{85.4} &74.3 & 76.9 & \textbf{80.3}\\
         & MCP-Universe \tiny{(Success Rate)}  & 46.5 & 47.9 & \textbf{50.7} & 35.6 &29.4 & \textbf{45.9} \\
      &  MCP-Mark {\tiny (Pass@1)} &33.3 & \textbf{50.9} & 43.1 &20.4 &24.4 & \textbf{38.0}\\
      & Tool-Decathlon {\tiny (Pass@1)}  & \textbf{38.6} & 29.0 &36.4 & 17.6& 16.0& \textbf{35.2} \\  
      
    \bottomrule
    \end{tabular}
    
    \label{tab:main}
\end{table}

\newmodel{} achieves similar performance with GPT-5-high on reasoning tasks, but is slightly worse than Gemini-3.0-Pro. Compared to K2-Thinking, \newmodel{} achieves comparable scores with substantially fewer output tokens, as shown in Table \ref{tab:long_model_performance}. These performance gains can be attributed to the increased computational resources allocated to RL training. Over recent months, we have observed consistent performance improvements correlating with extended RL training budget, which already exceeds 10\% of the pre-training cost. We hypothesize that reasoning capabilities could be further enhanced with additional computational budget allocation. Notably, the performance of \newmodel{} presented herein is constrained by a length constraint reward model; upon removal of the restriction, we observe further improvement in model performance, as detailed in Section \ref{longoutput}.

In code agent evaluations, \newmodel{} significantly outperforms open-source LLMs on both SWE-bench Verified and Terminal Bench 2.0, demonstrating its potential within real-world coding workflows. Regarding Terminal Bench 2.0, as previously noted, our context management strategy for the 'thinking mode' is currently incompatible with Terminus; consequently, the reported score of 46.4 was achieved using the Claude Code framework. We also evaluated \newmodel{} with Terminus in non-thinking mode, yielding a score of 39.3. For SWE-bench Verified, the primary score was obtained using our internal framework. Robustness tests across other settings—including the Claude Code and RooCode frameworks, as well as non-thinking mode—produced consistent results, ranging from 72 to 74.

For the search agent evaluation, we assess our models using a standard commercial search API. Since \newmodel{} supports a maximum context length of only 128K, approximately 20\%+ of the test cases exceed this limit. To address this, we employ a context management method to derive the final score. For reference, the score is 51.4 without context management. Further details are provided in Section \ref{sec:context}.

On tool-use benchmarks, \newmodel{} substantially narrows the performance gap between open-source and closed-source LLMs, though it remains below frontier models. For $\tau^2$-bench, we employ the model itself as the user agent, achieving final category scores of 63.8 (Airline), 81.1 (Retail), and 96.2 (Telecom). For the MCP benchmarks, we employ the function calling format and place tool outputs within messages designated with the 'tool' role, rather than the 'user' role.
During our testing, we observed that \newmodel{} frequently engages in redundant self-verification, generating excessively long trajectories. This tendency often causes the context length to exceed the 128K limit, particularly in tasks such as MCP-Mark GitHub and Playwright evaluation. Consequently, this phenomenon hinders the final performance of \newmodel{}. However, integrating context management strategies can further enhance performance. We identify this as a direction for future work and a practical consideration for users. Even if \newmodel{} suffers from the issue, it still significantly outperforms existing open models. 
Notably, since the environments and toolsets employed in these benchmarks were not encountered during RL training, the observed improvements demonstrate \newmodel's capacity to generalize its reasoning strategies to out-of-domain agentic scenarios. The evaluation of non-thinking model in the agent scenario is shown in Appendix Table \ref{tab:nonthink}.

\subsection{Results of \highmodel{}}
\label{longoutput}

\begin{table}[t]
    \centering
    % \small
    \caption{
    Benchmark performance and efficiency of reasoning models. For each benchmark, cells show accuracy and output token count (in thousands). The highest accuracy per benchmark is in bold; the second-highest is underlined.
    }
    \setlength{\tabcolsep}{4pt}
    \label{tab:long_model_performance}
    \resizebox{0.95\linewidth}{!}{%
    \begin{tabular}{@{}l llll|l@{}}
        \toprule
        \multirow{2}{*}{\textbf{Benchmark}} 
        & \textbf{GPT-5} & \textbf{Gemini-3.0} & \textbf{Kimi-K2} & \textbf{DeepSeek-V3.2} & \textbf{DeepSeek-V3.2} \\
        & \textbf{High} & \textbf{Pro} & \textbf{Thinking} & \textbf{Thinking} & \textbf{Speciale} \\
        \midrule
        {AIME 2025 {\tiny (Pass@1)}} 
        & 94.6 (13k) 
        & \underline{95.0} (15k) 
        & 94.5 (24k)
        & 93.1 (16k) 
        & \textbf{96.0} (23k) \\
        
        {HMMT Feb 2025 {\tiny (Pass@1)}} 
        & 88.3 (16k) 
        & \underline{97.5} (16k) 
        & 89.4 (31k)
        & 92.5 (19k) 
        & \textbf{99.2} (27k) \\
        
        {HMMT Nov 2025 {\tiny (Pass@1)}} 
        & 89.2 (20k) 
        & \underline{93.3} (15k) 
        & 89.2 (29k)
        & 90.2 (18k) 
        & \textbf{94.4} (25k) \\
        
        {IMOAnswerBench {\tiny (Pass@1)}} 
        & 76.0 (31k)
        & \underline{83.3} (18k)
        & 78.6 (37k)
        & 78.3 (27k)
        & \textbf{84.5} (45k) \\
        
        {LiveCodeBench {\tiny (Pass@1-COT)}} 
        & 84.5 (13k) 
        & \textbf{90.7} (13k) 
        & 82.6 (29k)
        & 83.3 (16k) 
        & \underline{88.7} (27k) \\
        
        {CodeForces {\tiny (Rating)}} 
        & 2537 (29k) 
        & \textbf{2708} (22k) 
        & -
        & {2386} (42k) 
        & \underline{2701} (77k) \\
        
        {GPQA Diamond {\tiny (Pass@1)}} 
        & \underline{85.7} (8k) 
        & \textbf{91.9} (8k)
        & 84.5 (12k)
        & 82.4 (7k)
        & \underline{85.7} (16k) \\
        
        {HLE {\tiny (Pass@1)}}
        & 26.3 (15k)
        & \textbf{37.7} (15k)
        & 23.9 (24k)
        & 25.1 (21k)
        & \underline{30.6} (35k) \\
        \bottomrule
    \end{tabular}
    }
\end{table}

Table \ref{tab:long_model_performance} demonstrates that \highmodel{} achieves superior performance by leveraging increased reasoning tokens, surpassing the state-of-the-art Gemini-3.0-Pro across multiple benchmarks.
Remarkably, as shown in Table \ref{tab:imo_ioi}, this general-purpose model attains gold-medal level performance in the 2025 International Olympiad in Informatics (IOI) and the ICPC World Finals (ICPC WF) without targeted training.
Furthermore, by incorporating techniques from \citet{deepseek-math-v2}, the model excels in complex proof tasks, reaching gold-medal thresholds in the 2025 International Mathematical Olympiad (IMO) and China Mathematical Olympiad (CMO)\footnote{We evaluated the English version of CMO 2025. The IMO 2025 and CMO 2025 problems, together with the inference code, can be found at: \url{https://github.com/deepseek-ai/DeepSeek-Math-V2}.}. Detailed evaluation protocols are provided in Appendix \ref{appendix:ioi_eval}.

However, the token efficiency of \highmodel{} remains significantly inferior to that of Gemini-3.0-Pro. 
To mitigate deployment costs and latency, we imposed stricter token constraints during the training of the official \newmodel{}, aiming to optimize the trade-off between performance and cost.
We believe that token efficiency remains a critical area for future investigation.

\begin{table}[htbp]
\centering
\caption{Performance of \highmodel{} in top-tier mathematics and coding competitions.
For ICPC WF 2025, we report the number of submissions for each successfully solved problem. \highmodel{} ranked 2nd in ICPC WF 2025 and 10th in IOI 2025.
}
\label{tab:imo_ioi}

\begin{tabular}{lcccccc cc}
\toprule
\textbf{Competition} & \textbf{P1} & \textbf{P2} & \textbf{P3} & \textbf{P4} & \textbf{P5} & \textbf{P6} & \textbf{Overall} & \textbf{Medal} \\
\midrule
IMO 2025 & 7 & 7 & 7 & 7 & 7 & 0 & 35/42 & \textcolor{gold}{Gold} \\
CMO 2025 & 18 & 18 & 9 & 21 & 18 & 18 & 102/126 & \textcolor{gold}{Gold} \\
IOI 2025 & 100 & 82 & 72 & 100 & 55 & 83 & 492/600 & \textcolor{gold}{Gold} \\
\bottomrule
\end{tabular}

\vspace{1em}

\setlength{\tabcolsep}{4pt}
% \resizebox{\linewidth}{!}{%
\begin{tabular}{lcccccccccccc cc}
\toprule
\textbf{Competition} & \textbf{A} & \textbf{B} & \textbf{C} & \textbf{D} & \textbf{E} & \textbf{F} & \textbf{G} & \textbf{H} & \textbf{I} & \textbf{J} & \textbf{K} & \textbf{L} & \textbf{Overall} & \textbf{Medal} \\
\midrule
ICPC WF 2025 & 3 & - & 1 & 1 & 2 & 2 & - & 1 & 1  & 1 & 1 & 1 & 10/12 & \textcolor{gold}{Gold} \\
\bottomrule
\end{tabular}
% }%
\end{table}

\subsection{Synthesis Agentic Tasks}
In this section, we perform ablation experiments to study the effect of synthetic agentic tasks. We focus on two questions. First, are synthetic tasks sufficiently challenging for reinforcement learning? Second, how well do these synthetic tasks generalize, i.e., can they transfer to different downstream tasks or real-world environments?

To address the first question, we randomly sample 50 instances from the general synthesized agentic tasks and evaluate both the model used for synthesis and frontier closed-source LLMs. As shown in Table~\ref{tab:synthesis-eval}, \newmodel-Exp attains an accuracy of only 12\%, while frontier closed-source models achieve at most 62\%. These results indicate that the synthetic data include agentic tasks that are challenging for both \newmodel-Exp and frontier closed-source models.
\begin{table}[h]
    \centering
    \footnotesize
    \caption{ \centering Accuracy of general synthesized tasks on different models.}
    \begin{tabular}{c | c  c c c }
    \toprule
     Pass@K& DeepSeek-v3.2-Exp & Sonnet-4.5  &Gemini-3.0 Pro & GPT-5-Thinking\\
    \midrule
    1& 12\%&34\%&51\%& 62\%  \\
    \midrule
    2& 18\% &47\%& 65\%&75\%  \\
    \midrule
    4& 26\%&62\%&74\% & 82\% \\
    \bottomrule
    \end{tabular}
    
    \label{tab:synthesis-eval}
\end{table}

\begin{figure}[t]
    \centering
    \includegraphics[width=0.95\linewidth]{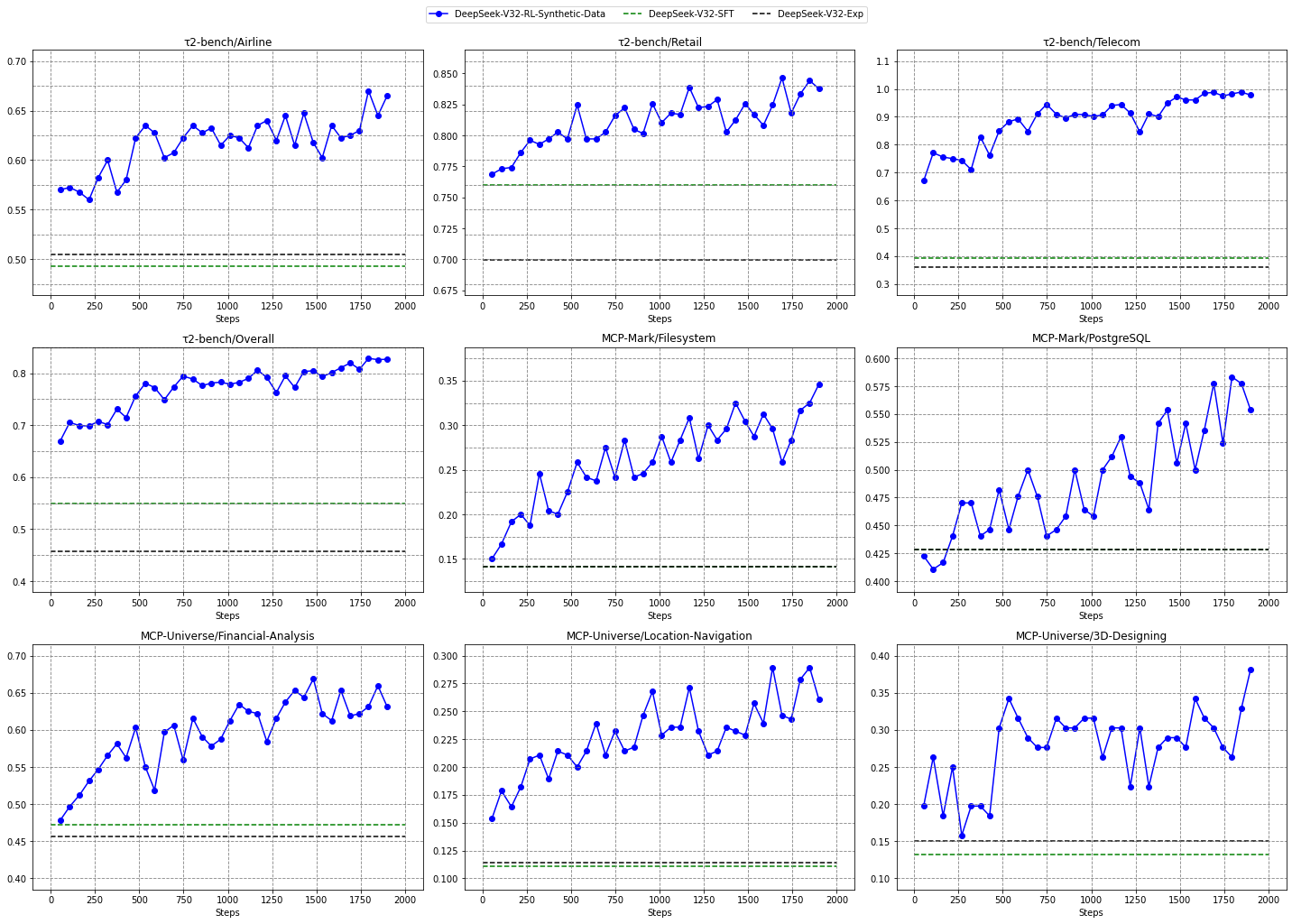}
    \caption{
       RL training of DeepSeek-V3.2-SFT using exclusively synthetic general agent data.
    }
    \label{fig:synthesis-rl}
\end{figure}

To investigate whether RL on synthetic data can generalize to different tasks or real-world environments, we apply RL to the SFT checkpoint of \newmodel~ (denoted \newmodel-SFT). To exclude the effects of long CoT and other RL data, we conduct RL only on synthetic agentic tasks in non-thinking mode. We then compare the model with \newmodel-SFT and \newmodel-Exp, where \newmodel-Exp is trained with RL only in search and code environments. As shown in Figure~\ref{fig:synthesis-rl}, large-scale RL on synthetic data yields substantial improvements over \newmodel-SFT on Tau2Bench, MCP-Mark, and MCP-Universe benchmarks. In contrast, restricting RL to code and search scenarios does not improve performance on these benchmarks, further highlighting the potential of synthetic data.

\subsection{Context Management of Search Agent}

\label{sec:context}
\begin{figure}[h]
    \centering
    \includegraphics[width=0.95\linewidth]{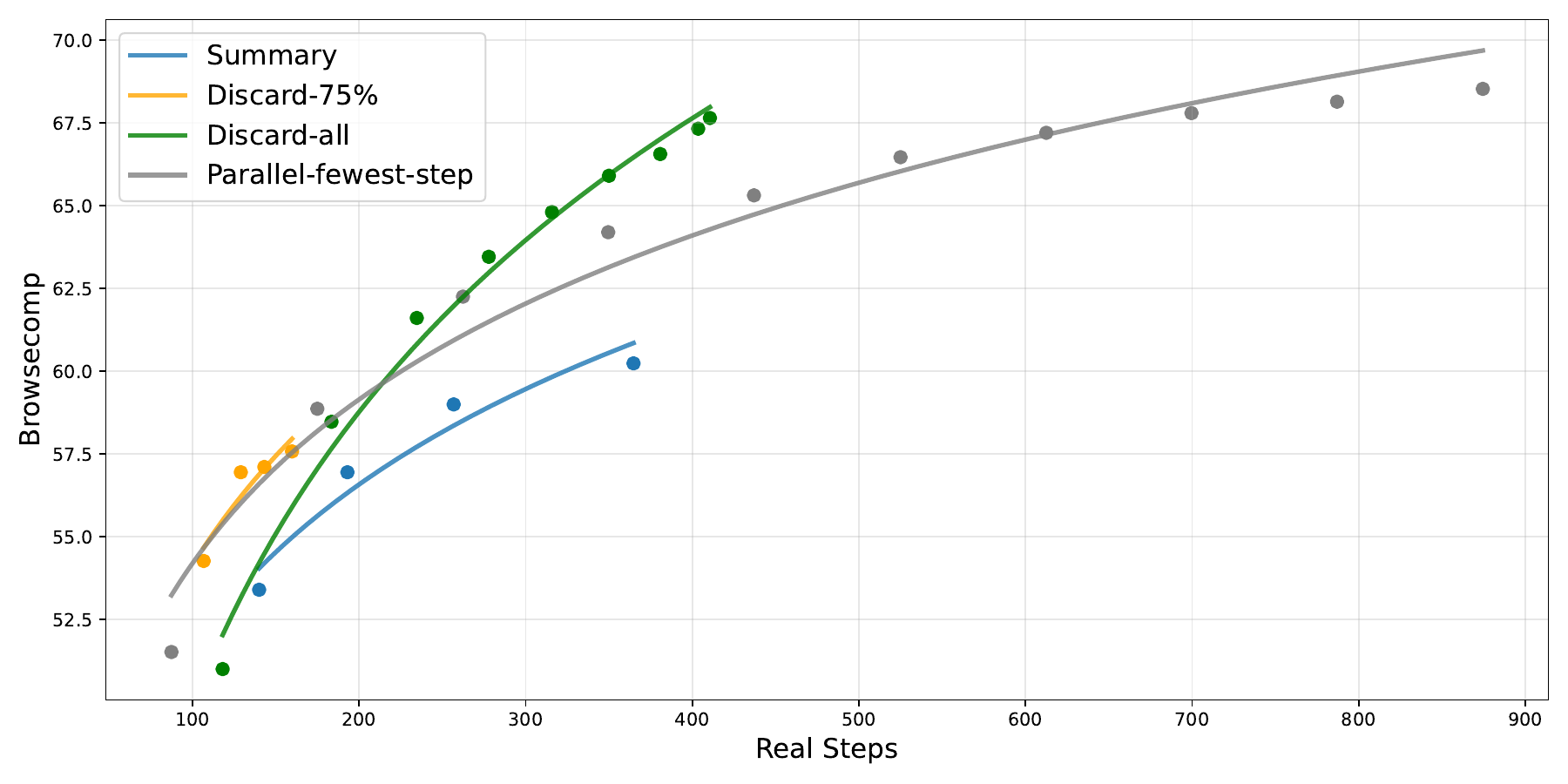}
    \caption{
    Accuracy of Browsecomp with different test-time compute expansion strategies.
    }
    \label{fig:search}
\end{figure}

Even with extended context windows such as 128k, agentic workflows, particularly in search-based scenarios, frequently encounter maximum length limitations that prematurely truncate the reasoning process. 
This bottleneck inhibits the full realization of test-time compute potential. 
To address this, we introduce context management employing simple strategies to extend token budgets at test time， when the token usage exceeds 80\% of the context window length. 
These strategies include 
(1) \textbf{Summary}, which summarizes the overflowed trajectory and re-initiates the rollout;
(2) \textbf{Discard-75\%}, which discards the first 75\% tool call history in the trajectory to free up spaces;
(3) \textbf{Discard-all}, which resets the context by discarding all previous tool call history (similar to the new context tool~\citep{opus4.5}). 
For comparison, we also implement a parallel scaling baseline, \textbf{Parallel-fewest-step}, which samples N independent trajectories and selects the trajectory with the fewest steps.

We evaluate these strategies on the BrowseComp benchmark~\citep{wei2025browsecomp}. 
As illustrated in Figure~\ref{fig:search}, under varying compute budgets, context management leads to significant performance gains by allowing the model to scale up test-time compute, providing more space to perform additional execution steps.
For example, Summary extends the average steps to 364, achieving a performance improvement of up to 60.2. However, its overall efficiency is relatively low.
Despite its simplicity, Discard-all performs well in both efficiency and scalability, achieving a score of 67.6, comparable to parallel scaling while using significantly fewer steps.

In summary, test-time compute can be scaled either serially through context management or in parallel, both effectively extending the model's problem-solving capacity.
However, different strategies exhibit varying efficiency and scalability. 
Thus, it is crucial to account for actual compute costs when benchmarking model performance. 
Meanwhile, finding the optimal combination of serial and parallel scaling to maximize both efficiency and scalability remains a crucial direction for future work.

\section{Conclusion, Limitation, and Future Work}
In this work, we introduced \newmodel, a framework that effectively bridges the gap between computational efficiency and advanced reasoning capabilities. Using DSA, we addressed critical computation complexity without sacrificing long-context performance. By increasing computational budget, \newmodel{} achieves comparable performance with GPT-5 on reasoning benchmarks. Finally, the integration of our large-scale agentic task synthesis pipeline significantly enhances tool-use proficiency, unlocking new possibilities for robust and generalizable AI agents with open LLM. Furthermore, our high-compute variant, \highmodel{}, validated by gold-medal achievements in the IMO and IOI, sets a milestone for open LLMs.

Despite these achievements, we acknowledge certain limitations when compared to frontier closed-source models such as Gemini-3.0-Pro. First, due to fewer total training FLOPs, the breadth of world knowledge in \newmodel{} still lags behind that of leading proprietary models. We plan to address this knowledge gap in future iterations by scaling up the pre-training compute. Second, token efficiency remains a challenge; \newmodel{} typically requires longer generation trajectories (i.e., more tokens) to match the output quality of models like Gemini-3.0-Pro. Future work will focus on optimizing the intelligence density of the model's reasoning chains to improve efficiency. Third, solving complex tasks is still inferior to frontier models, motivating us to further refine our foundation model and post-training recipe.
\bibliographystyle{abbrvnat}
\bibliography{main}

% \newpage
\appendix
\section*{Appendices}

\section{MHA and MQA Modes of MLA}
\label{appendix:mqa_mha}

\begin{figure}[h]
    \centering
    \subfigure[MHA mode of MLA.]{
        \includegraphics[width=0.475\textwidth]{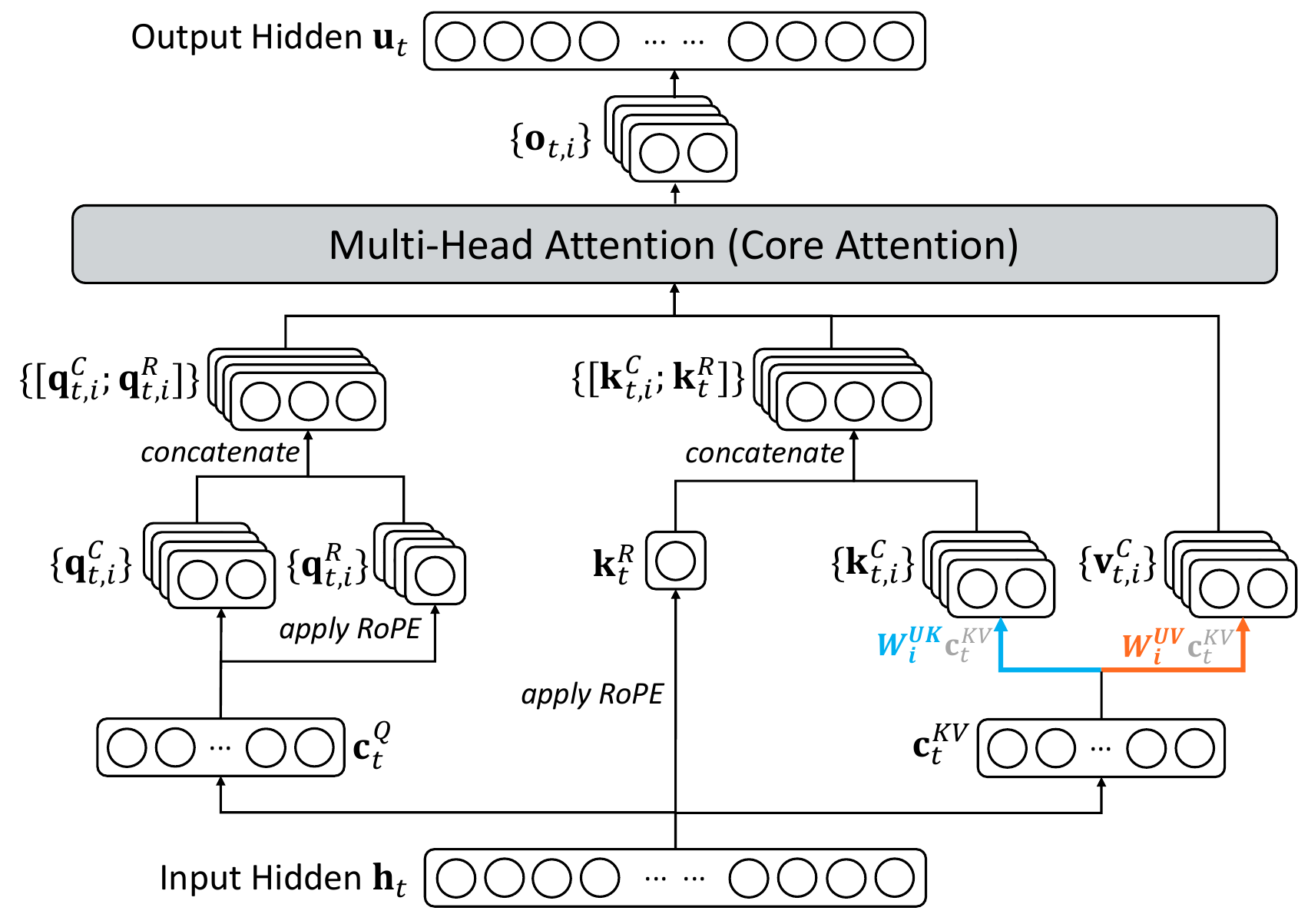}
    }
    \hspace{0.01cm}
    \subfigure[MQA mode of MLA.]{
        \includegraphics[width=0.475\textwidth]{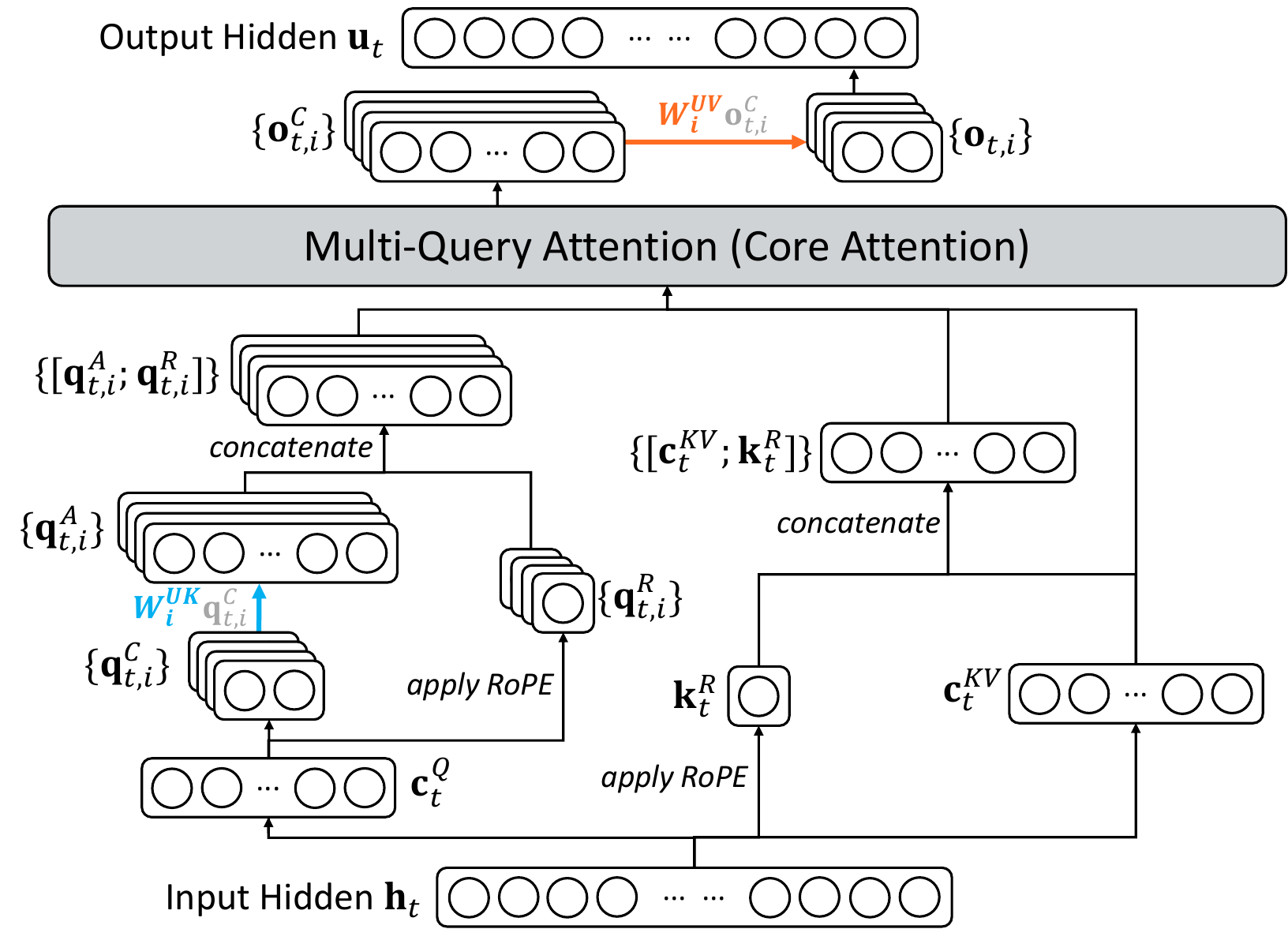}
    }
    \caption{
    Illustration of the MHA and MQA modes of MLA.
    For \oldmodel{}, the MHA mode is used for training and prefilling, while the MQA mode is used for decoding. 
    }
    \label{fig:mha_mqa_mode}
\end{figure}

Figure~\ref{fig:mha_mqa_mode} illustrates two aspects of MLA -- the MHA and MQA modes -- as well as the transformation between them.

% \section{Parity Evaluation of the Base Model}

% \begin{figure}[h]
%     \centering
%     \includegraphics[width=0.8\textwidth]{figures/arena.png}
%     \caption{Chatbot Arena Elo scores as of November 10, 2025, demonstrating that DeepSeek V3.2 Exp exhibits no regression in human preference benchmarks.}
% \end{figure}

%\newpage
\section{Cold Start Template}

\begin{table}[htbp]
\centering
\small
\begin{minipage}{\textwidth}
\centering    \caption{An example of the reasoning data system prompt. The system prompt requires the model to output the reasoning process in the tag <think></think>.}
    \label{tab:think_template}
 \begin{tabular}{p{0.1\textwidth}|p{0.77\textwidth}}
    \toprule

    Reasoning System Prompt & You are an expert Python programmer. You will be given a question (problem specification) and will generate a correct Python program that matches the specification and passes all tests. Please first reason before giving the final answer. 
    The reasoning process enclosed within <think> </think>. The final answer is output after the </think> tag.   \\ \hline
    Prompt &  Given a linked list, swap every two adjacent nodes and return its head ... \\ \hline
    Reasoning Response & <think>

    ...

    </think>

    [FINAL ANSWER]
    
    \\ 
     \bottomrule
    \end{tabular}

\end{minipage}
\hfill
\begin{minipage}{\textwidth}
\centering
   \small
        \caption{ \{TOOL-DESCRIPTIONS\} and \{TOOLCALL-FORMAT\} will be replaced with the specific tools and our designed toolcall format.}
    \label{tab:agent_template}
    
    \begin{tabular}{p{0.1\textwidth}|p{0.77\textwidth}}
    \toprule

    Agent System Prompt &  Use Python interpreter tool to execute Python code. The code will not be shown to the user. This tool should be used for internal reasoning, but not for code that is intended to be visible to the user (e.g. when creating plots, tables, or files). When you send a message containing Python code to python, it will be executed in a stateful Jupyter notebook environment. python will respond with the output of the execution or time out after 120.0 seconds.

    \#\# Tools
    
You have access to the following tools:

\{TOOL-DESCRIPTIONS\}

  Important: ALWAYS adhere to this exact format for tool use:

\{TOOLCALL-FORMAT\}

    \\ \hline
    Prompt &Given a linked list, swap every two adjacent nodes and return its head ... \\ \hline
    Agent Response & [MULTI-TURN TOOLCALL]
    
    [FINAL ANSWER] \\ 
     \bottomrule
    \end{tabular}

\end{minipage}

\begin{minipage}{\textwidth}
 \centering
    \small
        \caption{  The model executes tool calls in thinking process.}
   \label{tab:agentthink_template}
       \begin{tabular}{p{0.1\textwidth}|p{0.77\textwidth}}
    \toprule

    Reasoning Required Agent System Prompt & You are a helpful assistant with access to a Python interpreter.
    
- You may use the Python tool **multiple times** during your reasoning, a.k.a in <think></think>, with a maximum of 20 code executions.

- Call the Python tool early in your reasoning to aid in solving the task. Continue reasoning and invoking tools as needed until you reach the final answer. Once you have the answer, stop reasoning and present your solution using Markdown and LaTeX.

- Do NOT invoke any tools in your presented final solution steps.

- To improve efficiency and accuracy, you should prefer code execution over language-based reasoning whenever possible. Keep your reasoning succinct; let the code do the heavy lifting.

    \#\# Tools
    
You have access to the following tools:

\{TOOL-DESCRIPTIONS\}

  Important: ALWAYS adhere to this exact format for tool use:

\{TOOLCALL-FORMAT\}

\\ \hline
    Prompt & Given a linked list, swap every two adjacent nodes and return its head ... \\ \hline
    Agent Response with Thinking &
    <think> 
    
    [MULTI-TURN Thinking-Then-TOOLCALL]

</think>
    
    [FINAL ANSWER]  \\ 
     \bottomrule
    \end{tabular}

    \end{minipage}

\end{table}
\newpage
\section{Non-thinking \newmodel~ Agentic Evaluation }
\begin{table}[htbp]
    \centering
    \footnotesize
    \setlength{\tabcolsep}{1.9pt}
    \caption{ Comparison between \newmodel~ non-thinking and thinking modes. The terminal bench scores are evaluated with the Claude Code framework in the table. Non-thinking score of Terminal Bench 2.0 with Terminus framework is 39.3. 
      }
    \begin{tabular}{@{}c l | c  c  }
    \toprule
    & \multirow{1}{*}{\centering \textbf{Benchmark {\tiny (Metric)}}}  & \textbf{non-thinking}  & \textbf{thinking}
\\
    \midrule

     \multirow{3}{*}{Code Agent} & Terminal Bench 2.0 {\tiny (Acc)}  &37.1 & 46.4\\
      & SWE Verified {\tiny (Resolved)}  &72.1& 73.1\\
      & SWE Multilingual {\tiny (Resolved)}  &68.9 & 70.2 \\  \midrule

         \multirow{4}{*}{ToolUse}  &$\tau^2$-bench {\tiny (Pass@1)} & 77.2& 80.3 \\
         & MCP-Universe \tiny{(Success Rate)}  &38.6  & 45.9\\
      &  MCP-Mark {\tiny (Pass@1)} & 26.5 &38.0  \\
      & Tool-Decathlon {\tiny (Pass@1)}  & 25.6 & 35.2   \\  
      
    \bottomrule
    \end{tabular}
    
    \label{tab:nonthink}
\end{table}

The performance of non-thinking mode is slightly worse than the thinking mode, but still competitive. 

\section{Evaluation Method of IOI, ICPC World Final, IMO, and CMO}
\label{appendix:ioi_eval}
For all competitions, the model's maximum generation length is set to 128k. No tools or internet access are used, and testing strictly adheres to the contest's time and attempt limits.

For the IOI evaluation, we designed our submission strategy in accordance with the official competition rules, which permit up to 50 submissions per problem and score each submission based on the maximum points achieved across all subtasks. Specifically, we first sampled 500 candidate solutions for each problem, then applied a multi-stage filtering pipeline. In the initial stage, we eliminated invalid submissions that failed to pass the provided sample test cases or exceeded the length constraints. Subsequently, we employed the DeepSeek-V32-Exp model to identify and remove samples in which the model explicitly indicated an inability or refusal to solve the problem. From the remaining valid candidates, we selected the 50 samples with the longest thinking traces for final submission.

For the ICPC evaluation, we adapted the same filtering methodology but with a smaller initial sample size. We generated 32 candidate solutions per problem and applied the identical filtering criteria to select submissions.

In the IMO and CMO tasks, we employ a generate-verify-refine loop. The model iteratively improves its solution until it achieves a perfect self-evaluation or hits the maximum revision cap, identical to the process in \citet{deepseek-math-v2}.

\newpage
\section{Author List}
\noindent
\textbf{Research \& Engineering}: 
Aixin Liu,
Aoxue Mei,
Bangcai Lin,
Bing Xue,
Bingxuan Wang,
Bingzheng Xu,
Bochao Wu,
Bowei Zhang,
Chaofan Lin,
Chen Dong,
Chengda Lu,
Chenggang Zhao,
Chengqi Deng,
Chenhao Xu,
Chong Ruan*,
Damai Dai,
Daya Guo,
Dejian Yang,
Deli Chen,
Erhang Li,
Fangqi Zhou*,
Fangyun Lin,
Fucong Dai,
Guangbo Hao,
Guanting Chen,
Guowei Li,
H. Zhang,
Hanwei Xu,
Hao Li,
Haofen Liang,
Haoran Wei,
Haowei Zhang,
Haowen Luo,
Haozhe Ji,
Honghui Ding,
Hongxuan Tang,
Huanqi Cao,
Huazuo Gao,
Hui Qu,
Hui Zeng,
Jialiang Huang,
Jiashi Li,
Jiaxin Xu,
Jiewen Hu,
Jingchang Chen,
Jingting Xiang,
Jingyang Yuan,
Jingyuan Cheng,
Jinhua Zhu,
Jun Ran*,
Junguang Jiang,
Junjie Qiu,
Junlong Li*,
Junxiao Song,
Kai Dong,
Kaige Gao,
Kang Guan,
Kexin Huang*,
Kexing Zhou,
Kezhao Huang,
Kuai Yu,
Lean Wang,
Lecong Zhang,
Lei Wang,
Liang Zhao,
Liangsheng Yin*,
Lihua Guo,
Lingxiao Luo,
Linwang Ma,
Litong Wang,
Liyue Zhang,
M.S. Di,
M.Y Xu,
Mingchuan Zhang,
Minghua Zhang,
Minghui Tang,
Mingxu Zhou,
Panpan Huang,
Peixin Cong,
Peiyi Wang,
Qiancheng Wang,
Qihao Zhu,
Qingyang Li,
Qinyu Chen,
Qiushi Du,
Ruiling Xu,
Ruiqi Ge,
Ruisong Zhang,
Ruizhe Pan,
Runji Wang,
Runqiu Yin,
Runxin Xu,
Ruomeng Shen,
Ruoyu Zhang,
S.H. Liu,
Shanghao Lu,
Shangyan Zhou,
Shanhuang Chen,
Shaofei Cai,
Shaoyuan Chen,
Shengding Hu,
Shengyu Liu,
Shiqiang Hu,
Shirong Ma,
Shiyu Wang,
Shuiping Yu,
Shunfeng Zhou,
Shuting Pan,
Songyang Zhou,
Tao Ni,
Tao Yun,
Tian Pei,
Tian Ye,
Tianyuan Yue,
Wangding Zeng,
Wen Liu,
Wenfeng Liang,
Wenjie Pang,
Wenjing Luo,
Wenjun Gao,
Wentao Zhang,
Xi Gao,
Xiangwen Wang,
Xiao Bi,
Xiaodong Liu,
Xiaohan Wang,
Xiaokang Chen,
Xiaokang Zhang,
Xiaotao Nie,
Xin Cheng,
Xin Liu,
Xin Xie,
Xingchao Liu,
Xingkai Yu,
Xingyou Li,
Xinyu Yang,
Xinyuan Li*,
Xu Chen,
Xuecheng Su,
Xuehai Pan,
Xuheng Lin,
Xuwei Fu,
Y.Q. Wang,
Yang Zhang,
Yanhong Xu,
Yanru Ma,
Yao Li,
Yao Li,
Yao Zhao,
Yaofeng Sun,
Yaohui Wang,
Yi Qian,
Yi Yu,
Yichao Zhang,
Yifan Ding,
Yifan Shi,
Yiliang Xiong,
Ying He,
Ying Zhou,
Yinmin Zhong,
Yishi Piao,
Yisong Wang,
Yixiao Chen,
Yixuan Tan,
Yixuan Wei,
Yiyang Ma,
Yiyuan Liu,
Yonglun Yang,
Yongqiang Guo,
Yongtong Wu,
Yu Wu,
Yuan Cheng,
Yuan Ou,
Yuanfan Xu,
Yuduan Wang,
Yue Gong*,
Yuhan Wu,
Yuheng Zou,
Yukun Li,
Yunfan Xiong,
Yuxiang Luo,
Yuxiang You,
Yuxuan Liu,
Yuyang Zhou,
Z.F. Wu,
Z.Z. Ren,
Zehua Zhao,
Zehui Ren,
Zhangli Sha,
Zhe Fu,
Zhean Xu,
Zhenda Xie,
Zhengyan Zhang,
Zhewen Hao,
Zhibin Gou,
Zhicheng Ma,
Zhigang Yan,
Zhihong Shao,
Zhixian Huang,
Zhiyu Wu,
Zhuoshu Li,
Zhuping Zhang,
Zian Xu,
Zihao Wang,
Zihui Gu,
Zijia Zhu,
Zilin Li,
Zipeng Zhang,
Ziwei Xie,
Ziyi Gao,
Zizheng Pan,
Zongqing Yao

\noindent
\textbf{Data Annotation:}
Bei Feng,
Hui Li,
J.L. Cai,
Jiaqi Ni,
Lei Xu,
Meng Li,
Ning Tian,
R.J. Chen,
R.L. Jin,
S.S. Li,
Shuang Zhou,
Tianyu Sun,
X.Q. Li,
Xiangyue Jin,
Xiaojin Shen,
Xiaosha Chen,
Xinnan Song,
Xinyi Zhou,
Y.X. Zhu,
Yanping Huang,
Yaohui Li,
Yi Zheng,
Yuchen Zhu,
Yunxian Ma,
Zhen Huang,
Zhipeng Xu,
Zhongyu Zhang

\noindent
\textbf{Business \& Compliance:}
Dongjie Ji,
Jian Liang,
Jianzhong Guo,
Jin Chen,
Leyi Xia,
Miaojun Wang,
Mingming Li,
Peng Zhang,
Ruyi Chen,
Shangmian Sun,
Shaoqing Wu,
Shengfeng Ye,
T.Wang,
W.L. Xiao,
Wei An,
Xianzu Wang,
Xiaowen Sun,
Xiaoxiang Wang,
Ying Tang,
Yukun Zha,
Zekai Zhang,
Zhe Ju,
Zhen Zhang,
Zihua Qu

Authors are listed alphabetically by their first name. 
Names marked with * denote individuals who have departed from our team. 

\end{CJK*}
\end{document}